\documentclass{article}

\usepackage{arxiv}
\usepackage[T1]{fontenc}
\usepackage[utf8]{inputenc}
\usepackage{graphicx}
\usepackage{placeins} 
\usepackage{amsmath,amssymb,mathtools}
\usepackage{siunitx}
\usepackage{microtype}
\usepackage[colorlinks=true,linkcolor=blue,citecolor=blue,urlcolor=blue]{hyperref}
\usepackage[style=numeric-comp,sorting=none,backend=biber,doi=true,isbn=false]{biblatex}
\title{Learning-Based Interpolation of Detector Gaps in Photon-Counting X-ray Imaging}

\author{
  Johannes B. Thalhammer$^{1,2,3}$ \quad
  Patrick Ilg$^{1, 2}$ \quad
  Tina Dorosti$^{1,2,3}$ \quad
  Sebastian Peterhansl$^{1,2}$ \\
  \bfseries Theresa Hiu$^{1,2}$ \quad
  Daniel Frey$^{1,2}$ \quad
  Zijin Huang$^{1,2}$ \\
  \bfseries Florian Schaff$^{1,2}$ \quad
  Daniela Pfeiffer$^{1,2,3}$ \quad
  Franz Pfeiffer$^{1,2,3}$ \\[1.5ex]
  \parbox{0.95\textwidth}{\centering\normalfont\small
    $^{1}$Chair of Biomedical Physics, Department of Physics, School of Natural Sciences, Technical University of Munich (TUM), Germany \\
    $^{2}$Munich Institute of Biomedical Engineering, TUM, Germany \\
    $^{3}$Department of Diagnostic and Interventional Radiology, School of Medicine and Health, Klinikum rechts der Isar, TUM, Germany \\[1ex]
    Corresponding author: \texttt{johannes.thalhammer@tum.de}}
}

\renewcommand{\shorttitle}{Learning-Based Detector-Gap Interpolation}
\hypersetup{
  pdftitle={Learning-Based Interpolation of Detector Gaps in Photon-Counting X-ray Imaging},
  pdfauthor={Johannes B. Thalhammer, Patrick Ilg, Tina Dorosti, Sebastian Peterhansl, Theresa Hiu, Daniel Frey, Zijin Huang, Florian Schaff, Daniela Pfeiffer, Franz Pfeiffer},
  pdfkeywords={detector gap interpolation, image inpainting, photon-counting detectors, GAN, Pix2Pix, computed tomography},
}

\begin{document}
\maketitle

\begin{abstract}
Image detectors assembled from tiled modules have physical gaps in the field of view that must be filled before downstream processing such as phase retrieval or tomographic reconstruction. Unlike natural-image inpainting, detector-gap interpolation must leave all other pixels unmodified and preserve the intensity statistics of the filled region, since even a small shift in mean intensity can cause ring artifacts after reconstruction. We adapted two generative models, a Wasserstein GAN with gradient penalty (WGAN-GP) and a conditional Pix2Pix model, to this task. We implemented a masking strategy that restricts all changes to the gap pixels and loss terms that constrain the mean intensity and the intensity distribution of the filled region. Both models were trained on nanoCT projections with simulated gaps and compared with linear interpolation, Navier--Stokes inpainting, and the deep image-inpainting network MADF, quantitatively on simulated gaps and visually on real detector gaps. WGAN-GP and Pix2Pix achieved the best perceptual similarity (LPIPS) and histogram distance and, apart from linear interpolation, the shortest inference time. Pix2Pix required about five times fewer training GPU-hours than WGAN-GP. Ablation studies showed that masking the network output contributes by far the most to plausible inpainting results. Applied to sinograms, Pix2Pix produced the most natural gap filling but improved the reconstruction only slightly compared with linear interpolation. The source code and the weights of the final Pix2Pix model are available at \url{https://github.com/J-3TO/detectorGapRemoval_Pix2Pix}.
\end{abstract}


\section{Introduction}
Photon-counting detectors offer several advantages over conventional detectors, but are commonly designed with individual detector tiles arranged into a larger panel, introducing physical gaps in the field of view. These gaps do not provide imaging information and can cause errors in subsequent processing steps, which propagate into artifacts in the reconstructed tomographic images.

Several classical algorithms exist to fill image gaps. Linear interpolation is fast but transmits only smooth, low-frequency information into the gap and fails to reproduce the surrounding texture. More sophisticated inpainting techniques, such as the fast-marching method of Telea~\cite{telainpainting} or the Navier--Stokes-based approach~\cite{NSInpainting}, better preserve local structure, but are slower, require manual parameter tuning, and still struggle to reproduce realistic texture.

Specifically for detector gaps, Brun et al. proposed a physics-informed approach: the masked measurement is reconstructed iteratively and forward-projected, and the forward-projected values are combined with partial-differential-equation-based inpainting to fill the gap~\cite{Brun_2018}. This approach is also computationally expensive, does not yield a fully realistic inpainting, and does not entirely remove ring artifacts.

For natural images, deep-learning-based inpainting has shown promising results. Examples are the cascaded refinement network with mask awareness (MADF) by Zhu et al.~\cite{MADFPaper} and LaMa by Suvorov et al., which uses Fourier convolutions to inpaint large gaps~\cite{suvorov2021resolution}. Diffusion models have also shown potential for inpainting, but their many sampling steps make them computationally expensive~\cite{latDiffusionModels, Lugmayr_2022_CVPR}.

Detector-gap interpolation also demands higher fidelity than natural-image inpainting, for which a plausible-looking result is sufficient. In detector data, even a small, visually imperceptible bias in the gap can cause artifacts downstream: a shift in the mean intensity of the filled region, for instance, leads to an incorrect thickness estimate in Paganin phase retrieval and may lead to artifacts after CT reconstruction. At the same time, pixels outside the gap must remain unmodified, as any change there corrupts valid measurement data.

The goal of this work was therefore to find inpainting methods that produce plausible results, reduce artifacts in subsequent processing such as phase retrieval and reconstruction, and are fast enough to be used within a reconstruction pipeline. MADF was used both in its original form pretrained by Zhu et al. and trained from scratch on simulated detector gaps, owing to its strong performance on small masks reported by Suvorov et al.~\cite{suvorov2021resolution}. In addition, two generative models, a Wasserstein GAN with gradient penalty (WGAN-GP)~\cite{gulrajani2017improvedtrainingwassersteingans} and Pix2Pix~\cite{isola2018imagetoimagetranslationconditionaladversarial}, were implemented and trained for the task, again on simulated detector gaps. A masking strategy ensured that only pixels within the gap were modified, and dedicated loss terms preserved the intensity statistics of the inpainted region.

\section{Methods}

\subsection{Dataset}

The data were acquired on a nanoCT setup comprising an Excillum NanoTube~N1 \SI{60}{\kilo\volt} Beta transmission-target source and a DECTRIS Pilatus3~2M photon-counting detector with a \SI{1}{\milli\meter}-thick silicon sensor \cite{nanoCTConferenceAbstract}. The detector consists of 24 modules ($3 \times 8$) of $487 \times 195$ pixels each (pixel size \SI{172}{\micro\meter}), which form an array of $1475 \times 1679$ pixels (first column of each panel in Figure~\ref{fig:resultGapInterpol}). Between the modules, two 7-pixel gaps run horizontally and seven 17-pixel gaps run vertically. The detector is mounted with a fixed horizontal offset of about a quarter of the module width. This decorrelates the gap positions from the rotation axis, so that the vertical gaps do not occlude the same object feature throughout a scan. Motorized stages allow the source-to-object distance to be set from \SIrange{1.5}{50}{\milli\meter} and the source-to-detector distance from \SIrange{420}{1920}{\milli\meter}.

To compare architectures and configurations, a smaller dataset containing only projection images was used (projection dataset). It comprises 12,552 projections from 25 scans of specimens including soft tissue, bone, and cell aggregates, each image covering one detector module of $487 \times 195$ pixels. The scans were split into training (15 scans, 7,368 images), validation (5 scans, 2,520 images), and test (5 scans, 2,664 images) sets. To increase robustness, natural photographs from the ImageNet ILSVRC2012 validation split~\cite{Russakovsky2015} were mixed into the training data at a rate of 30\%.
For each training step, a $256 \times 128$-pixel patch was randomly cropped from a nanoCT or ImageNet image and normalized to zero mean and unit variance. A synthetic gap (horizontal, vertical, or a combination of both) with a random width of 7--40~pixels, matching the physical gap widths of our detector 7 and 17~pixels, was masked out. The masked patch was passed to the network together with the binary gap mask as a second input channel. Random Gaussian blur and Poisson noise served as further augmentation.

A second, larger dataset (extended dataset) comprises the projection images and sinograms of 70 additional scans acquired on the same setup. The scans were split into training (47 scans; 88,949 projections, 68,667 sinograms), validation (11 scans; 23,585 projections, 16,071 sinograms), and test (12 scans; 23,667 projections, 17,532 sinograms) sets. As in the projection dataset, sinogram patches were cropped within a single detector module and normalized to zero mean and unit variance. In a sinogram, a missing detector column appears as a straight vertical line across all projection angles; synthetic gaps in sinograms were therefore restricted to vertical stripes of 7--40~pixels width. Gaussian blur and Poisson noise were applied as for the projection dataset.

\subsection{Loss terms}
\label{sec:lossterms}

Let $y$ denote a normalized ground-truth patch, $\hat{y}$ the corresponding prediction, and $\Omega$ the set of gap pixels (where the binary mask $m=0$). Three loss terms were used. The first is the mean-squared error (MSE) within the gap,
\begin{equation}
    \mathcal{L}_{\mathrm{mse}} = \frac{1}{|\Omega|}\sum_{i \in \Omega} \left(\hat{y}_i - y_i\right)^2 .
    \label{eq:mseloss}
\end{equation}
The second, the mean gap difference, is the absolute difference between the mean predicted and the mean ground-truth intensity in the gap,
\begin{equation}
    \mathcal{L}_{\mathrm{mean}} \;=\; \left| \, \frac{1}{|\Omega|}\sum_{i \in \Omega} \hat{y}_i \; - \; \frac{1}{|\Omega|}\sum_{i \in \Omega} y_i \, \right| .
    \label{eq:gapmeandiff}
\end{equation}
The third term constrains the shape of the intensity distribution in the gap and is inspired by the histogram losses of~\cite{adl2022, deephist2020}. The pixel values in $\Omega$ are binned into $n_b = 128$ bins of width $\Delta = 6\sigma_\Omega / n_b$ spanning $[\mu_\Omega - 3\sigma_\Omega,\; \mu_\Omega + 3\sigma_\Omega]$, where $\mu_\Omega$ and $\sigma_\Omega$ are the mean and standard deviation of the ground truth in $\Omega$. This gives the bin edges $e_k = \mu_\Omega - 3\sigma_\Omega + k\Delta$, $k = 0,\dots,n_b$.

\begin{figure}[tbp]
    \centering
    \includegraphics[width=1.0\linewidth]{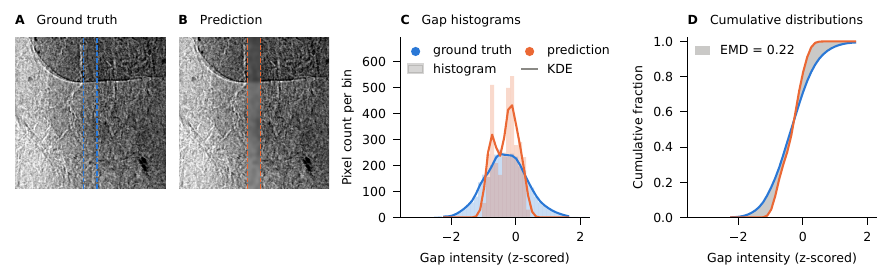}
    \caption{\textbf{Histogram-based training loss.} \textbf{A} Ground-truth patch with a detector gap (dashed lines). \textbf{B} Gap prediction with correct mean intensity yet over-smoothed.
    \textbf{C} Hard and soft histograms of the gap pixels. In the soft histogram, each pixel contributes a Gaussian of one bin width. \textbf{D} Cumulative distributions of the soft histograms. The area between them is the estimated Earth Mover's Distance (EMD).}
    \label{fig:methodHistoLoss}
\end{figure}

A hard histogram is piecewise constant in the pixel values, so its gradient vanishes almost everywhere. Therefore, a soft histogram is used: each pixel value is represented by a Gaussian kernel of width $s$, and each bin receives the fraction of the kernel's area that falls inside it. For $v \in \{y, \hat{y}\}$, the $k$-th bin is
\begin{equation}
    \tilde{h}_k(v) = \sum_{i \in \Omega} \left[ \Phi\!\left(\frac{e_k - v_i}{s}\right) - \Phi\!\left(\frac{e_{k-1} - v_i}{s}\right) \right],
    \qquad k = 1,\dots,n_b,
    \label{eq:softhist}
\end{equation}
where $\Phi$ is the cumulative distribution function of the standard normal distribution, and the kernel width was set to one bin, $s = \Delta$. Following~\cite{deephist2020}, the soft histograms are compared with the Earth Mover's Distance (EMD), i.e.\ the area between the two cumulative distributions $C_k(v) = \frac{1}{|\Omega|} \sum_{j=1}^{k} \tilde{h}_j(v)$:
\begin{equation}
    \mathcal{L}_{\mathrm{EMD}} = \Delta \sum_{k=1}^{n_b} \left| C_k(\hat{y}) - C_k(y) \right| .
    \label{eq:emdloss}
\end{equation}
Gaps with a nearly constant ground truth ($\sigma_\Omega < 10^{-4}$) are excluded, as they define no meaningful distribution. Figure~\ref{fig:methodHistoLoss} illustrates the hard and soft histograms and the EMD.

The WGAN-GP and Pix2Pix generators were trained with a weighted sum of these three terms and an adversarial term $\mathcal{L}_{\mathrm{adv}}$:
\begin{equation}
    \mathcal{L}_G =
        \lambda_{\mathrm{adv}}\,\mathcal{L}_{\mathrm{adv}}
        + \lambda_{\mathrm{mse}}\,\mathcal{L}_{\mathrm{mse}}
        + \lambda_{\mathrm{mean}}\,\mathcal{L}_{\mathrm{mean}}
        + \lambda_{\mathrm{EMD}}\,\mathcal{L}_{\mathrm{EMD}}.
    \label{eq:generator_loss}
\end{equation}

\subsection{Network architectures and training}
\label{sec:architectures}

\begin{figure}[tbp]
    \centering
    \includegraphics[width=1.0\linewidth]{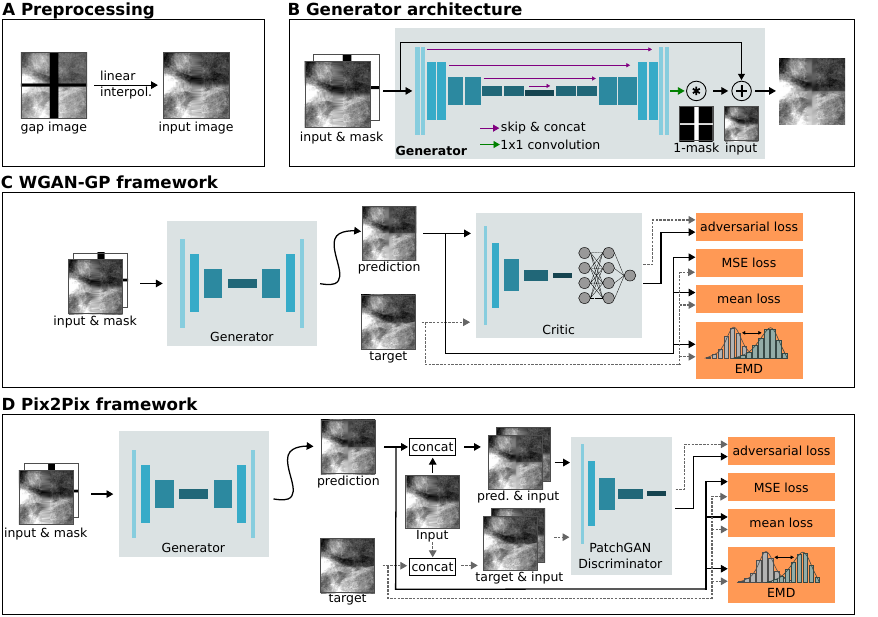}
    \caption{Overview of the detector-gap interpolation pipeline. \textbf{A} Preprocessing: the gaps are filled by linear interpolation. \textbf{B} Generator: a U-Net receives the pre-interpolated image and the binary gap mask. Its output is multiplied by the inverted mask ($1-\text{mask}$) and added to the input, so that only the gap pixels are modified and all other pixels pass through unchanged. \textbf{C} WGAN-GP framework~\cite{gulrajani2017improvedtrainingwassersteingans}: the critic scores the generator prediction and the target image, and its output is used for calculating the adversarial loss. The MSE, mean, and Earth Mover's Distance (EMD) losses (Section~\ref{sec:lossterms}) are computed between prediction and target. \textbf{D} Pix2Pix framework~\cite{isola2018imagetoimagetranslationconditionaladversarial}: same losses as in \textbf{C}, but the PatchGAN discriminator receives the input paired with either the prediction or the target.}
    \label{fig:methodDetectorGapInterpol}
\end{figure}

Figure~\ref{fig:methodDetectorGapInterpol} gives an overview of the pipeline. WGAN-GP and Pix2Pix share the same generator, a five-level U-Net~\cite{Ronneberger2015} (encoder channel widths $64$--$128$--$256$--$512$--$1024$) with bilinear upsampling. Its input is the linearly pre-interpolated image, concatenated with the binary gap mask. The predicted residual is multiplied by the inverted mask, which sets it to zero outside the gap, and added to the input. The generator thus learns a correction to the linear interpolation inside the gap, while all other pixels pass through unchanged.

In the WGAN-GP approach (Figure~\ref{fig:methodDetectorGapInterpol} \textbf{C}), the generator is trained against a critic based on EfficientNetV2-S~\cite{TanEffNet}, in which batch normalization was replaced by layer normalization, as required for the gradient penalty. The critic scores gap-free images against images inpainted by the generator and is trained with the WGAN-GP objective~\cite{gulrajani2017improvedtrainingwassersteingans}, with a gradient-penalty weight of $\lambda_{\mathrm{gp}}=10$ and $n_{\mathrm{critic}}=5$ critic updates per generator update. In the generator loss (Eq.~\eqref{eq:generator_loss}), the adversarial term is the negative mean critic score of the inpainted images, $\mathcal{L}_{\mathrm{adv}} = -\mathbb{E}\!\left[D\!\left(G(x)\right)\right]$, where $x$ denotes the masked, pre-interpolated input, $G$ the generator, and $D$ the critic.

In the Pix2Pix approach (Figure~\ref{fig:methodDetectorGapInterpol} \textbf{D}), the generator is trained in a conditional adversarial setting~\cite{isola2018imagetoimagetranslationconditionaladversarial}. A $70\times70$ PatchGAN discriminator receives the pre-interpolated input paired with either the prediction or the ground truth and outputs a map of real/fake scores per patch. It is trained with a binary cross-entropy loss with one-sided label smoothing ($0.9$). For the generator, $\mathcal{L}_{\mathrm{adv}}$ is the standard conditional GAN loss, $\mathcal{L}_{\mathrm{adv}} = -\mathbb{E}\!\left[\log D(x, G(x))\right]$, i.e.\ the binary cross-entropy between the discriminator's patch scores for the prediction and the label ``real''. Both generators were optimized with Adam~\cite{Kingma2015Adam:Optimization} ($\mathrm{lr}=2\times10^{-4}$, $\beta_1=0.5$, $\beta_2=0.999$).

Adversarial and pixel-wise losses alone can permit small systematic shifts in the mean or distribution of the gap intensities. Such shifts are visually negligible but bias further quantitative processing, e.g. Paganin phase retrieval~\cite{Paganin2002SimultaneousObject} in our case. $\mathcal{L}_{\mathrm{mean}}$ and $\mathcal{L}_{\mathrm{EMD}}$ were added to suppress them. The loss weights were tuned heuristically so that all terms have a similar order of magnitude: $\lambda_{\mathrm{adv}}=1$ (WGAN-GP) or $2$ (Pix2Pix), $\lambda_{\mathrm{mse}}=5$, $\lambda_{\mathrm{mean}}=100$, and $\lambda_{\mathrm{EMD}}=1{,}000$.

WGAN-GP and Pix2Pix were each trained for 700 epochs on a single NVIDIA RTX 3090; MADF was trained for 3,000 epochs on three RTX 3090 GPUs.

To enable inpainting in the sinogram domain, the Pix2Pix model was then trained further on a mixture of both datasets (projection and extended dataset). Each sample in a mini-batch was drawn with probability 0.5 from the sinograms of the extended dataset and otherwise from the projection dataset (including its 30\% ImageNet share), resulting in 50\% sinogram, 35\% projection, and 15\% ImageNet patches. The epoch length was kept at that of the projection dataset (7,368 samples), and the validation set was mixed in the same proportions. Generator and discriminator were initialized with the weights of the projection-only model, and the optimizer state was reset; all other settings were unchanged. This joint model was trained for 1,600 additional epochs on a single RTX 3090.

\subsection{Comparison methods}

WGAN-GP and Pix2Pix were compared with MADF~\cite{MADFPaper}, both pretrained on the Places dataset~\cite{placesDataset} and trained from scratch on the nanoCT data (with 30\% ImageNet images, as above), with Navier--Stokes inpainting as implemented in OpenCV~\cite{NSInpainting, opencv_library}, and with linear interpolation.

\subsection{Metrics}
\label{sec:metrics}
Three metrics were computed on simulated gaps in $256\times128$-pixel patches from the test set of the projection dataset. The mean gap difference $\mathcal{L}_{\mathrm{mean}}$ (Eq.~\eqref{eq:gapmeandiff}) measures a systematic intensity bias in the gap.

The histogram distance compares the shape of the intensity distribution in the gap. Hard histograms $h_k$ and $\hat{h}_k$ of the ground-truth and predicted gap pixels, binned as in Section~\ref{sec:lossterms}, are compared with a log-cosh distance adapted from Ghahremani et al.~\cite{adl2022},
\begin{equation}
    d_{\mathrm{hist}} = \frac{1}{n_b\,HW} \sum_{k=1}^{n_b} \log\cosh\!\left(h_k - \hat{h}_k\right),
    \label{eq:histdist}
\end{equation}
normalized by the number of bins and the patch size $HW$. The log-cosh function is quadratic for small and linear for large differences, which makes the distance smooth yet robust to a few strongly mismatched bins.

Finally, the Learned Perceptual Image Patch Similarity (LPIPS)~\cite{zhang2018lpips} with an AlexNet backbone measures perceptual and textural quality. It is computed over the full patch. For LPIPS, each normalized patch was clipped at $\mu\pm3\sigma$, mapped to $[0,1]$, and replicated across three channels.

\subsection{Inference}
All models were applied to $128\times128$-pixel patches with an overlap of 16 pixels, in which the predictions were linearly blended.
For the reconstructions, all samples were processed with an identical pipeline in which only the inpainting step differed: the projections were flat-field corrected, normalized for primary beam fluctuations, corrected pixel-wise for beam hardening (linearization with an automatically optimized blend of aluminum and copper), and corrected for defective pixels. The gaps were then inpainted in transmission, either in the projections or in the sinogram domain; in the latter case, only vertical gaps were corrected in sinogram space and a central slab between horizontal gaps was reconstructed. After drift correction, the projections were converted to line integrals, phase retrieved with the Paganin method~\cite{Paganin2002SimultaneousObject} ($\delta/\mu = 1.5\times10^{-6}\,\mathrm{mm}$; spheroid and aorta only), corrected with a ring removal filter, and reconstructed with the Feldkamp--Davis--Kress (FDK) algorithm~\cite{Feldkamp1984}.

\FloatBarrier
\section{Results}

First, all methods are compared for inpainting in the projection domain. Ablation studies of WGAN-GP and Pix2Pix then show how the individual design choices (masking, linear pre-interpolation, and the EMD and mean loss terms) affect the result. Lastly, the final Pix2Pix model is applied to sinograms, and its effect on the reconstructions is shown.

\subsection{Method comparison}
\begin{figure}[tbp]
    \centering
    \includegraphics[width=1.0\linewidth]{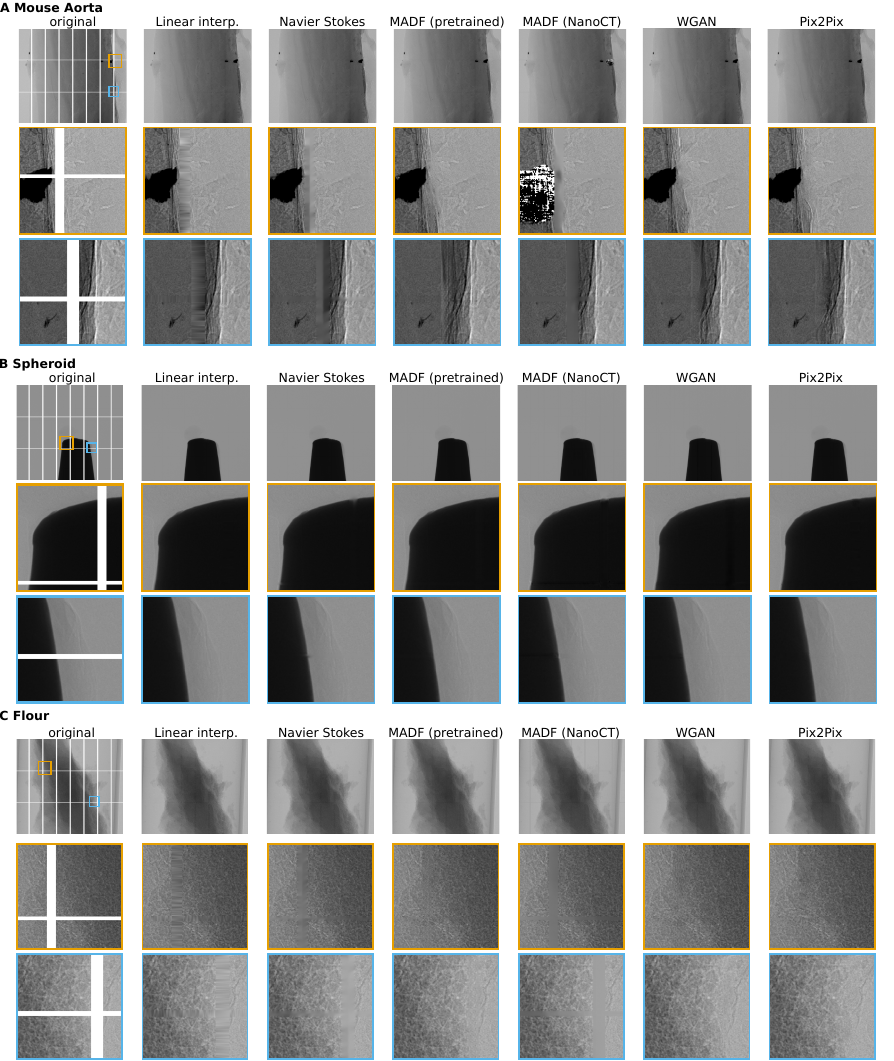}
    \caption{Inpainting results of all methods for \textbf{A} a mouse aorta, \textbf{B} a spheroid, and \textbf{C} a flour sample. Full image: $1475\times1679$ pixels; zoom regions: $200\times200$ and $150\times150$ pixels.}
    \label{fig:resultGapInterpol}
\end{figure}

Figure~\ref{fig:resultGapInterpol} shows measured projections with real detector gaps and the inpainting results of all methods for three samples that were not part of the training data: \textbf{A} a mouse aorta, \textbf{B} a spheroid, and \textbf{C} a flour sample. Linear interpolation works well for the spheroid, which shows high contrast between sample holder and background and little noise, but performs worse on noisier or more textured projections (\textbf{A}, \textbf{C}). Navier--Stokes inpainting yields smooth results not replicating the texture of the surrounding area and does not preserve the edge in \textbf{A}. MADF pretrained on natural images matches the overall intensities well despite never having seen nanoCT data, but produces an embroidery-like structure in the gaps. MADF trained on nanoCT data yields smoother results but alters the black area in \textbf{A}, as MADF network is not constrained to the gap region. WGAN-GP and Pix2Pix produce similar results that match both the intensities and the structure of the image area surrounding the gap.

\begin{figure}[tbp]
    \centering
    \includegraphics[width=1.0\linewidth]{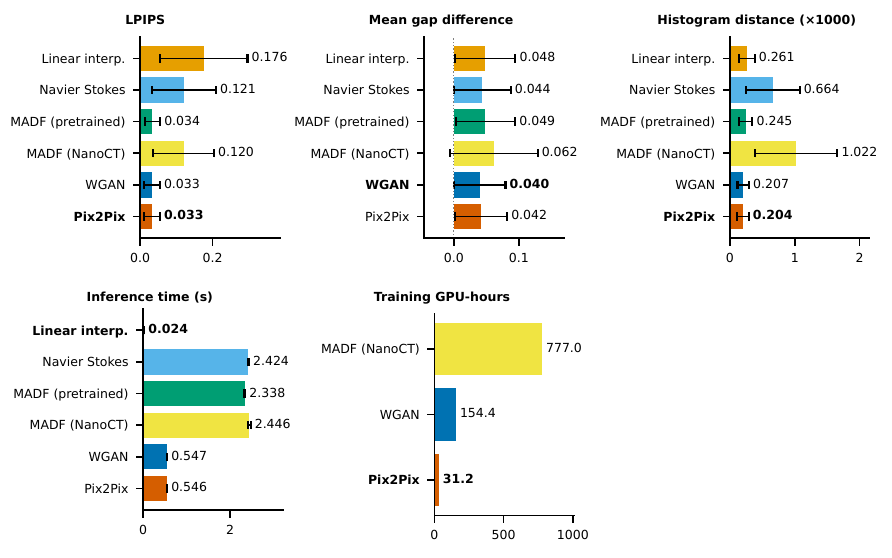}
    \caption{Quantitative comparison of all methods on simulated gaps in the test set. Top: LPIPS, mean gap difference, and histogram distance ($\times1000$). Bottom: inference time per $256\times128$-pixel patch and training time in GPU-hours on an NVIDIA RTX 3090. For all quantities, lower is better; the best value in each panel is shown in bold. Bars show the mean and error bars the standard deviation, except for the training time, which is a single value per model.}
    \label{fig:metricGapInterpol}
\end{figure}

The metrics on simulated gaps (Figure~\ref{fig:metricGapInterpol}) agree with these observations. WGAN-GP and Pix2Pix achieve the best LPIPS and histogram distance and, apart from linear interpolation, the shortest inference time per patch. The pretrained MADF outperforms MADF trained on nanoCT data in all three metrics. Presumably, the nanoCT dataset was too small to train MADF from scratch, or the original training settings~\cite{MADFPaper}, including the VGG-based perceptual loss with ImageNet-pretrained weights, were not suited to this kind of data. Pix2Pix was also the fastest to train (31.2 GPU-hours, vs.\ 154.4 for WGAN-GP and 777.0 for MADF). All metrics show large standard deviations, which reflect the variation between patches of the diverse test set.

\subsection{WGAN-GP ablation study}

\begin{figure}[tbp]
    \centering
    \includegraphics[width=\linewidth,height=0.85\textheight,keepaspectratio]{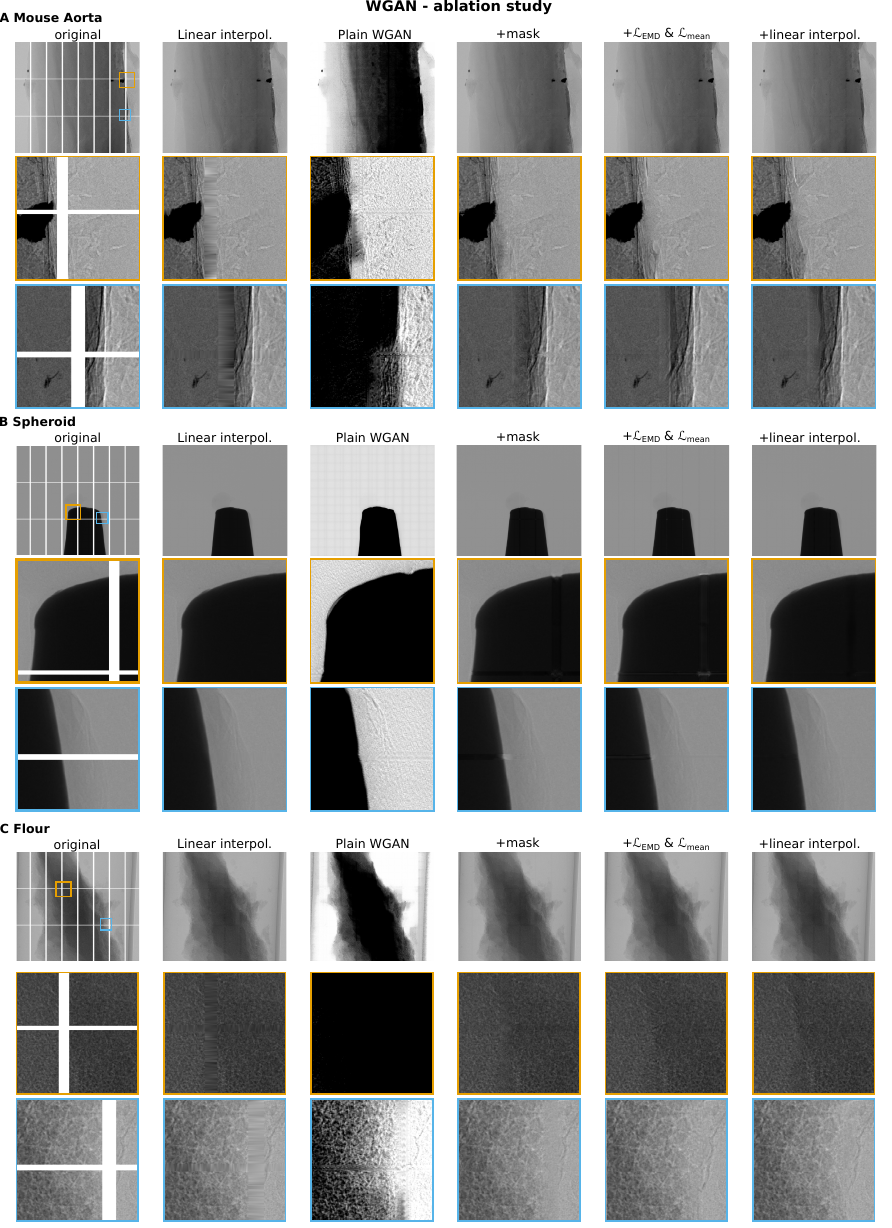}
    \caption{WGAN-GP ablation study on measured projections with real detector gaps of \textbf{A} a mouse aorta, \textbf{B} a spheroid, and \textbf{C} a flour sample. Columns: original projection, linear interpolation, plain WGAN-GP, and WGAN-GP with components added cumulatively from left to right (masking, $\mathcal{L}_{\mathrm{EMD}}$ and $\mathcal{L}_{\mathrm{mean}}$, linear pre-interpolation). Rows: full image ($1475\times1679$ pixels) and the regions marked in orange ($200\times200$ pixels) and blue ($150\times150$ pixels).}
    \label{fig:WGANAblation}
\end{figure}

Figure~\ref{fig:WGANAblation} shows visual examples of the WGAN-GP ablation study, and Figure~\ref{fig:MetricsInpaintingAblation} \textbf{A} the corresponding metrics. Without masking and with the gap pixels set to zero in the input and only the adversarial and MSE loss (plain WGAN-GP), the predicted pixel values are shifted. Multiplying the prediction with the inverted mask (+mask) stabilizes the output, but a slight offset in the mean of the inpainted pixels remains. Adding $\mathcal{L}_{\mathrm{EMD}}$ and $\mathcal{L}_{\mathrm{mean}}$ on top of the masking improves all three metrics and roughly halves the mean gap difference, yet the gap still stands out slightly from its surroundings (Figure~\ref{fig:WGANAblation} \textbf{B}). Linear pre-interpolation improves the LPIPS and the histogram distance slightly further and makes the gap less visible.

\subsection{Pix2Pix ablation study}

\begin{figure}[tbp]
    \centering
    \includegraphics[width=1.0\linewidth]{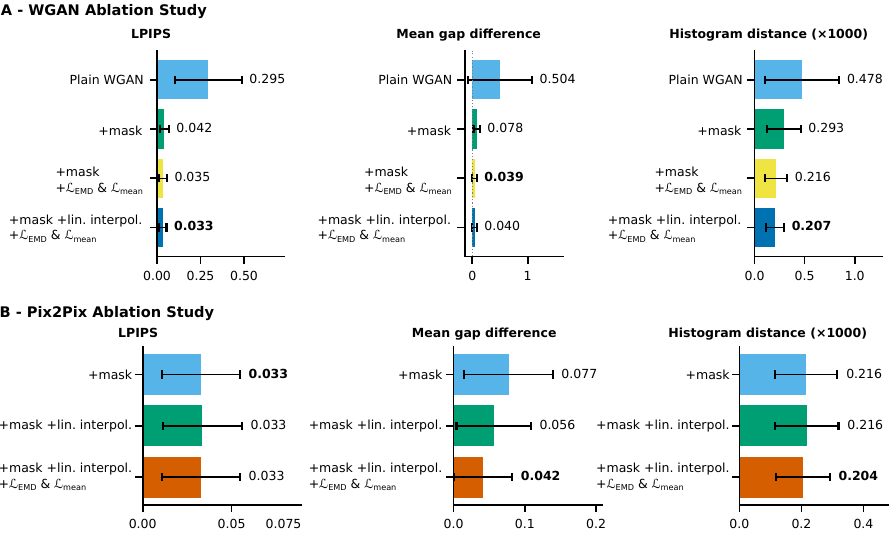}
    \caption{Metrics of the \textbf{A} WGAN-GP and \textbf{B} Pix2Pix ablation studies on simulated gaps in the test set: LPIPS, mean gap difference, and histogram distance ($\times1000$). From top to bottom, components are added cumulatively, from the baseline (plain WGAN-GP or masking only) to the final model (masking, linear pre-interpolation, $\mathcal{L}_{\mathrm{EMD}}$ and $\mathcal{L}_{\mathrm{mean}}$). For all metrics, lower is better. Bars show the mean and error bars the standard deviation; note the different axis scales in \textbf{A} and \textbf{B}.}
    \label{fig:MetricsInpaintingAblation}
\end{figure}

\begin{figure}[tbp]
    \centering
    \includegraphics[width=\linewidth,height=0.85\textheight,keepaspectratio]{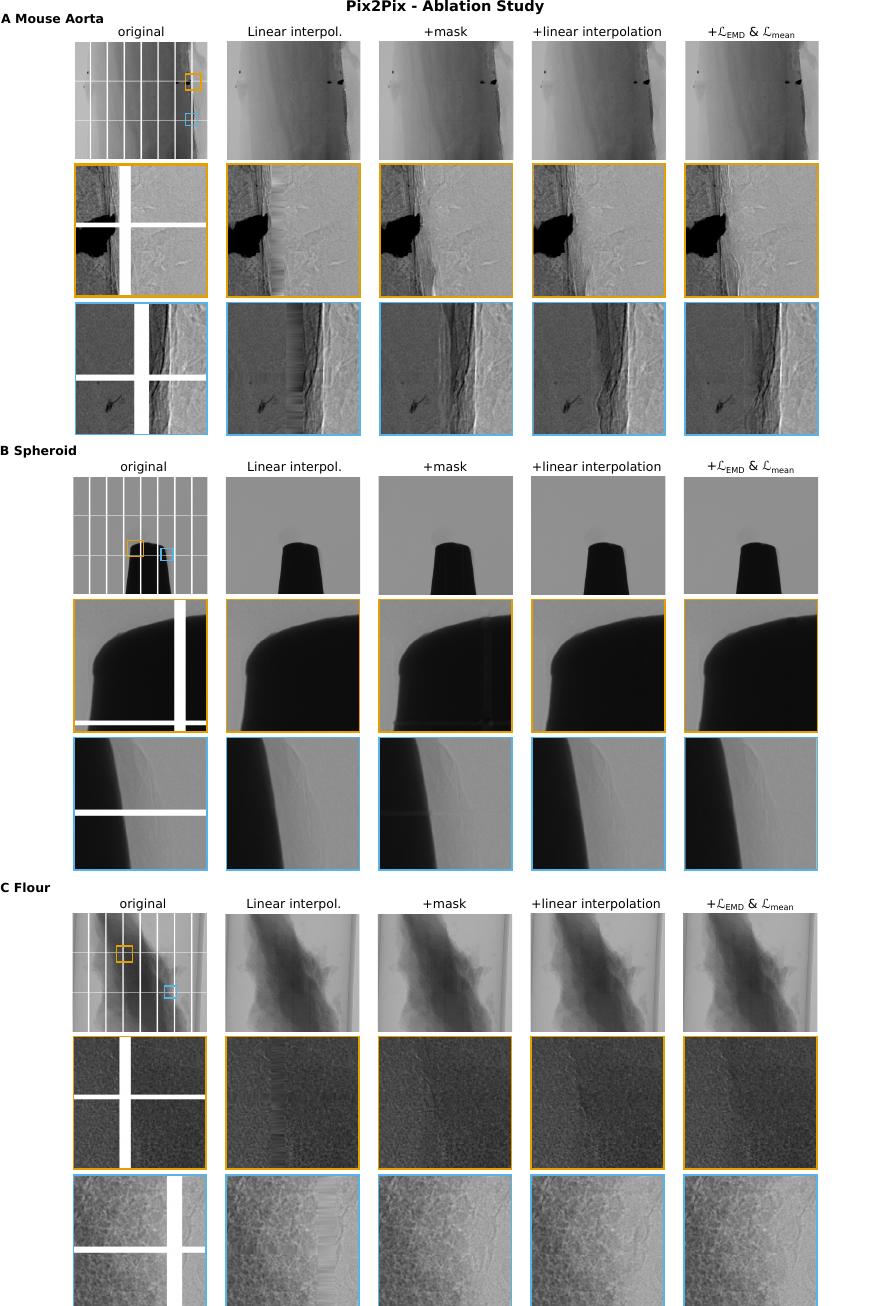}
    \caption{Pix2Pix ablation study on measured projections with real detector gaps of \textbf{A} a mouse aorta, \textbf{B} a spheroid, and \textbf{C} a flour sample. Columns: input with the gaps set to zero, linear interpolation, and Pix2Pix with components added cumulatively from left to right (masking, linear pre-interpolation, $\mathcal{L}_{\mathrm{EMD}}$ and $\mathcal{L}_{\mathrm{mean}}$). Rows: full image ($1475\times1679$ pixels) and the regions marked in orange ($200\times200$ pixels) and blue ($150\times150$ pixels).}
    \label{fig:Pix2PixAblation}
\end{figure}

Figure~\ref{fig:MetricsInpaintingAblation} \textbf{B} and Figure~\ref{fig:Pix2PixAblation} show the corresponding ablation study for Pix2Pix. Here, masking was always applied. With masking alone (+mask), a small mismatch in pixel values is visible in Figure~\ref{fig:Pix2PixAblation} \textbf{B}, which linear pre-interpolation reduces. Adding $\mathcal{L}_{\mathrm{EMD}}$ and $\mathcal{L}_{\mathrm{mean}}$ changes the results only slightly visually, but improves the mean gap difference and the histogram distance.

\subsection{Sinogram interpolation}

\begin{figure}[tbp]
    \centering
    \includegraphics[width=\linewidth,height=0.85\textheight,keepaspectratio]{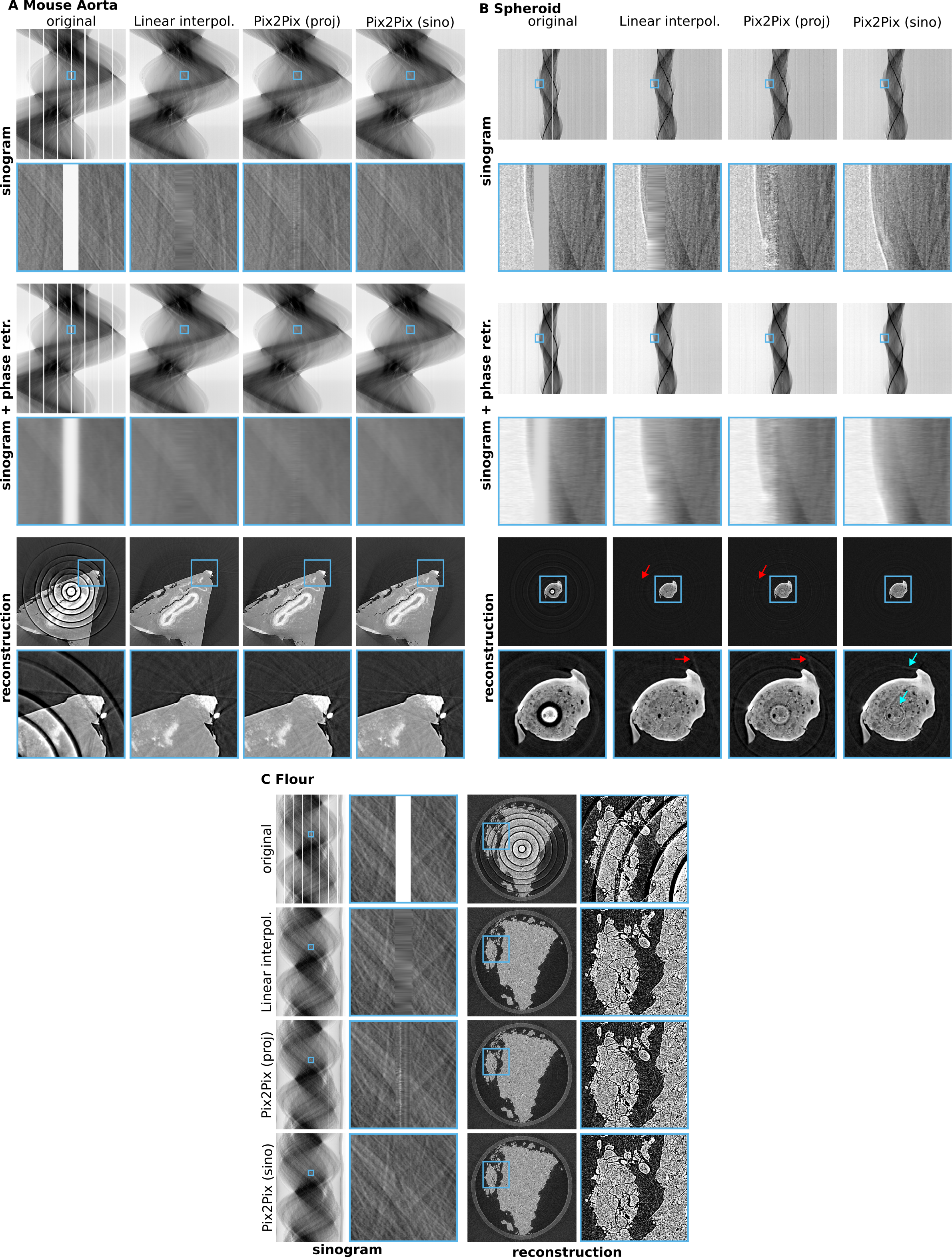}
    \caption{Effect of the gap filling on the sinogram, the Paganin phase-retrieved sinogram, and the FDK reconstruction (top to bottom, each with a zoomed-in region) for \textbf{A} the mouse aorta and \textbf{B} the spheroid. Columns: constant-filled gap (original), linear interpolation, Pix2Pix applied to the projections (proj), and Pix2Pix applied to the sinogram (sino). \textbf{C} Sinogram and reconstruction of the flour sample, with the four variants arranged in rows. Red arrows in \textbf{B} mark the streak artifact after linear interpolation and Pix2Pix (proj); cyan arrows mark the finer streak and the faint ring after Pix2Pix (sino).}
    \label{fig:SinogramInterpol}
\end{figure}

Figure~\ref{fig:SinogramInterpol} shows the effect of the gap filling on the reconstruction for the same three samples. For each sample, the central sinogram, the Paganin phase-retrieved sinogram (\textbf{A}, \textbf{B} only), and the FDK reconstruction are shown for four variants: the gap filled with a constant value of 1 (original), linear interpolation, and the final Pix2Pix model applied to the individual projections (proj) or directly to the sinogram (sino). No phase retrieval was applied to the flour sample, as its particles provide sufficient absorption contrast against the surrounding air, and the low-pass character of Paganin filtering would blur the fine granular structure.

In the sinograms, linear interpolation leaves clearly visible stripes. Pix2Pix (proj) also produces stripes, because each projection is inpainted independently and the filled values do not match between neighboring projection angles. Pix2Pix (sino) yields a gap that is almost indistinguishable from its surroundings. Phase retrieval, which acts as a low-pass filter, weakens these stripes but does not remove them completely.

In the reconstructions, the constant-filled gap causes severe ring artifacts in all three samples. All three inpainting methods reduce these artifacts substantially. In the aorta (\textbf{A}), the streaks left by linear interpolation are weaker with Pix2Pix (sino), while Pix2Pix (proj) shows more pronounced ring artifacts. In the spheroid (\textbf{B}), linear interpolation and Pix2Pix (proj) leave a streak artifact; Pix2Pix (sino) removes it but introduces a finer horizontal streak and a faint ring near the center. In the flour sample (\textbf{C}), the zoomed-in reconstructions of the three methods are very similar; only Pix2Pix (proj) shows faint rings near the center.

\FloatBarrier
\section{Conclusion}

Two generative models, WGAN-GP and Pix2Pix, were adapted for detector-gap inpainting and compared with MADF, Navier--Stokes inpainting, and linear interpolation. Both models achieved the best LPIPS and histogram distance with comparable scores and, apart from linear interpolation, the shortest inference time. Pix2Pix, however, required about five times fewer training GPU-hours than WGAN-GP (31.2 vs.\ 154.4). In the ablation studies, masking the network output had by far the largest effect, while linear pre-interpolation and the mean and EMD loss terms brought further, smaller improvements. Applied to sinograms, Pix2Pix produced the most natural gap filling, but the reconstructions showed no clear improvement over linear interpolation.

The results also suggest that visually realistic 2D inpainting alone does not guarantee artifact-free reconstructions. Inpainting in projection space is inconsistent across projection angles, and inpainting in sinogram space treats each detector row separately. Models with 3D context could enforce consistency in both directions and further improve the inpainting. Moreover, the models were trained on gaps simulated within detector modules and therefore do not capture effects that occur at real module edges, which could be a subject of future research. Nevertheless, the presented approach provides fast and visually convincing inpainting of photon-counting detector gaps and can serve as a basis for future work or as part of a more sophisticated preprocessing pipeline.

\section*{Acknowledgments}

We thank Niklas Bogovic and Philipp Kreiner (Department of Surgery, University Hospital Regensburg, Germany) for providing the liver spheroids, and Prof. Marcus R. Makowski, Avan Kader, and Jannis Marchand (Department of Diagnostic and Interventional Radiology, Technical University of Munich, Germany) for providing the mouse aorta specimens.

\printbibliography

@article{telainpainting,
author = {Telea, Alexandru},
year = {2004},
month = {01},
pages = {23-34},
title = {An Image Inpainting Technique Based on the Fast Marching Method},
volume = {9},
journal = {Journal of Graphics Tools},
doi = {10.1080/10867651.2004.10487596}
}

@INPROCEEDINGS{NSInpainting,
  author={Bertalmio, M. and Bertozzi, A.L. and Sapiro, G.},
  booktitle={Proceedings of the 2001 IEEE Computer Society Conference on Computer Vision and Pattern Recognition. CVPR 2001},
  title={Navier-stokes, fluid dynamics, and image and video inpainting},
  year={2001},
  volume={1},
  number={},
  pages={I-I},
  doi={10.1109/CVPR.2001.990497}}

@ARTICLE{MADFPaper,
  author={Zhu, Manyu and He, Dongliang and Li, Xin and Li, Chao and Li, Fu and Liu, Xiao and Ding, Errui and Zhang, Zhaoxiang},
  journal={IEEE Transactions on Image Processing},
  title={Image Inpainting by End-to-End Cascaded Refinement With Mask Awareness},
  year={2021},
  volume={30},
  number={},
  pages={4855-4866},
  doi={10.1109/TIP.2021.3076310}}

@article{suvorov2021resolution,
  title={Resolution-robust Large Mask Inpainting with Fourier Convolutions},
  author={Suvorov, Roman and Logacheva, Elizaveta and Mashikhin, Anton and Remizova, Anastasia and Ashukha, Arsenii and Silvestrov, Aleksei and Kong, Naejin and Goka, Harshith and Park, Kiwoong and Lempitsky, Victor},
  journal={arXiv preprint arXiv:2109.07161},
  year={2021}
}

@InProceedings{Lugmayr_2022_CVPR,
    author    = {Lugmayr, Andreas and Danelljan, Martin and Romero, Andres and Yu, Fisher and Timofte, Radu and Van Gool, Luc},
    title     = {RePaint: Inpainting Using Denoising Diffusion Probabilistic Models},
    booktitle = {Proceedings of the IEEE/CVF Conference on Computer Vision and Pattern Recognition (CVPR)},
    month     = {June},
    year      = {2022},
    pages     = {11461-11471}
}

@article{Brun_2018,
doi = {10.1088/1361-6501/aa91ad},
url = {https://doi.org/10.1088/1361-6501/aa91ad},
year = {2017},
month = {dec},
publisher = {IOP Publishing},
volume = {29},
number = {1},
pages = {014001},
author = {Brun, F and Delogu, P and Longo, R and Dreossi, D and Rigon, L},
title = {Inpainting approaches to fill in detector gaps in phase contrast computed tomography},
journal = {Measurement Science and Technology}
}

@article{opencv_library,
    author = {Bradski, G.},
    journal = {Dr. Dobb's Journal of Software Tools},
    title = {{The OpenCV Library}},
    year = {2000}
}

@inbook{Ronneberger2015,
  title = {U-Net: Convolutional Networks for Biomedical Image Segmentation},
  ISBN = {9783319245744},
  ISSN = {1611-3349},
  url = {http://dx.doi.org/10.1007/978-3-319-24574-4_28},
  DOI = {10.1007/978-3-319-24574-4_28},
  booktitle = {Medical Image Computing and Computer-Assisted Intervention – MICCAI 2015},
  publisher = {Springer International Publishing},
  author = {Ronneberger,  Olaf and Fischer,  Philipp and Brox,  Thomas},
  year = {2015},
  pages = {234–241}
}

@misc{gulrajani2017improvedtrainingwassersteingans,
      title={Improved Training of Wasserstein GANs},
      author={Ishaan Gulrajani and Faruk Ahmed and Martin Arjovsky and Vincent Dumoulin and Aaron Courville},
      year={2017},
      eprint={1704.00028},
      archivePrefix={arXiv},
      primaryClass={cs.LG},
      url={https://arxiv.org/abs/1704.00028},
}

@misc{isola2018imagetoimagetranslationconditionaladversarial,
      title={Image-to-Image Translation with Conditional Adversarial Networks},
      author={Phillip Isola and Jun-Yan Zhu and Tinghui Zhou and Alexei A. Efros},
      year={2018},
      eprint={1611.07004},
      archivePrefix={arXiv},
      primaryClass={cs.CV},
      url={https://arxiv.org/abs/1611.07004},
}

@inproceedings{Kingma2015Adam:Optimization,
    title = {{Adam: A method for stochastic optimization}},
    year = {2015},
    booktitle = {3rd International Conference on Learning Representations, ICLR 2015 - Conference Track Proceedings},
    author = {Kingma, Diederik P. and Ba, Jimmy Lei}
}

@article{Paganin2002SimultaneousObject,
    title = {{Simultaneous phase and amplitude extraction from a single defocused image of a homogeneous object}},
    year = {2002},
    journal = {Journal of Microscopy},
    author = {Paganin, D. and Mayo, S. C. and Gureyev, T. E. and Miller, P. R. and Wilkins, S. W.},
    number = {1},
    volume = {206},
    doi = {10.1046/j.1365-2818.2002.01010.x},
    issn = {00222720}
}

@INPROCEEDINGS {latDiffusionModels,
author = { Rombach, Robin and Blattmann, Andreas and Lorenz, Dominik and Esser, Patrick and Ommer, Bjorn },
booktitle = { 2022 IEEE/CVF Conference on Computer Vision and Pattern Recognition (CVPR) },
title = {{ High-Resolution Image Synthesis with Latent Diffusion Models }},
year = {2022},
volume = {},
ISSN = {},
pages = {10674-10685},
doi = {10.1109/CVPR52688.2022.01042},
url = {https://doi.ieeecomputersociety.org/10.1109/CVPR52688.2022.01042},
publisher = {IEEE Computer Society},
address = {Los Alamitos, CA, USA},
month =Jun}

@misc{adl2022,
  author        = {Ghahremani, Morteza and Khateri, Mohammad and Sierra, Alejandra and Tohka, Jussi},
  title         = {Adversarial Distortion Learning for Medical Image Denoising},
  year          = {2022},
  eprint        = {2204.14100},
  archivePrefix = {arXiv},
  primaryClass  = {eess.IV},
  url           = {https://arxiv.org/abs/2204.14100},
}

@inproceedings{zhang2018lpips,
  author    = {Zhang, Richard and Isola, Phillip and Efros, Alexei A. and Shechtman, Eli and Wang, Oliver},
  title     = {The Unreasonable Effectiveness of Deep Features as a Perceptual Metric},
  booktitle = {Proceedings of the IEEE/CVF Conference on Computer Vision and Pattern Recognition (CVPR)},
  year      = {2018},
  pages     = {586--595},
  doi       = {10.1109/CVPR.2018.00068},
}

@ARTICLE{deephist2020,
  title     = "Differentiable histogram loss functions for intensity-based
               image-to-image translation",
  author    = "Avi-Aharon, Mor and Arbelle, Assaf and Raviv, Tammy Riklin",
  journal   = "IEEE Trans. Pattern Anal. Mach. Intell.",
  publisher = "Institute of Electrical and Electronics Engineers (IEEE)",
  volume    =  45,
  number    =  10,
  pages     = "11642--11653",
  month     =  oct,
  year      =  2023
}

@article{TanEffNet,
  author       = {Mingxing Tan and
                  Quoc V. Le},
  title        = {EfficientNetV2: Smaller Models and Faster Training},
  journal      = {CoRR},
  volume       = {abs/2104.00298},
  year         = {2021},
  url          = {https://arxiv.org/abs/2104.00298},
  eprinttype   = {arXiv},
  eprint       = {2104.00298}
}

@article{Feldkamp1984,
  author  = {Feldkamp, L. A. and Davis, L. C. and Kress, J. W.},
  title   = {Practical cone-beam algorithm},
  journal = {Journal of the Optical Society of America A},
  year    = {1984},
  volume  = {1},
  number  = {6},
  pages   = {612--619},
  doi     = {10.1364/JOSAA.1.000612}
}

@ARTICLE{placesDataset,
  author={Zhou, Bolei and Lapedriza, Agata and Khosla, Aditya and Oliva, Aude and Torralba, Antonio},
  journal={IEEE Transactions on Pattern Analysis and Machine Intelligence}, 
  title={Places: A 10 Million Image Database for Scene Recognition}, 
  year={2018},
  volume={40},
  number={6},
  pages={1452-1464},
  doi={10.1109/TPAMI.2017.2723009}}

@inproceedings{nanoCTConferenceAbstract,
author = {Patrick Ilg and Oliver Schurius and Martin Dierolf and Benedikt G{\"u}nther and Franz Pfeiffer},
title = {{Improvements of a laboratory phase-contrast nano-computed tomography system}},
volume = {13152},
booktitle = {Developments in X-Ray Tomography XV},
editor = {Bert M{\"u}ller and Ge Wang},
organization = {International Society for Optics and Photonics},
publisher = {SPIE},
pages = {131521U},
year = {2024},
doi = {10.1117/12.3029536},
URL = {https://doi.org/10.1117/12.3029536}
}

@article{Russakovsky2015,
  title     = {{ImageNet} Large Scale Visual Recognition Challenge},
  author    = {Russakovsky, Olga and Deng, Jia and Su, Hao and Krause, Jonathan and Satheesh, Sanjeev and Ma, Sean and Huang, Zhiheng and Karpathy, Andrej and Khosla, Aditya and Bernstein, Michael and Berg, Alexander C. and Fei-Fei, Li},
  journal   = {International Journal of Computer Vision},
  volume    = {115},
  number    = {3},
  pages     = {211--252},
  year      = {2015},
  doi       = {10.1007/s11263-015-0816-y}
}

\end{document}